\DocumentMetadata{%
	pdfstandard=A-3u,
	pdfversion=1.7,
	lang=en-US,
}

\documentclass[balance,colorlinks,nofoot,upint,subscriptcorrection,varvw,mathalfa=cal=boondoxo,spanish,german,vietnamese,russian,greek]{asmeconf}

\usepackage{url}
\usepackage{hyperref}
\usepackage{graphicx}
\usepackage{natbib}
\usepackage[export]{adjustbox}
\usepackage{comment}
\usepackage{color}
\usepackage{xurl}
\usepackage{mathtools}
\usepackage{caption}
\usepackage{subcaption}
\usepackage{amsmath}
\usepackage{amsfonts}
\usepackage{algorithmic}
\usepackage{pifont}
\usepackage{textcomp}
\usepackage{multirow}
\usepackage{balance}

\hypersetup{%
	pdfauthor={John H. Lienhard},									  
	pdftitle={ASME Conference Paper LaTeX Template},                  
	pdfkeywords={ASME conference paper, LaTeX template, BibTeX style},
	pdfsubject = {Describes the asmeconf LaTeX template},			  
	pdflicenseurl={https://ctan.org/pkg/asmeconf},
}

\begin{document}




\title{Digital Twins in the Cloud: A Modular, Scalable and Interoperable Framework\\ for Accelerating Verification and Validation of Autonomous Driving Solutions}
 
%
%
%

\SetAuthors{%
Tanmay Samak$^{\star,}$\affil{}\JointFirstAuthor\CorrespondingAuthor{\href{mailto:tsamak@clemson.edu}{tsamak@clemson.edu}},
Chinmay Samak$^{\star,}$\affil{}\JointFirstAuthor,
Giovanni Martino$^{\star}$\affil{},
Pranav Nair$^{\star}$\affil{},
Venkat Krovi$^{\star}$\affil{}
}

\SetAffiliation{}{$^{\star}$Clemson University International Center for Automotive Research (CU-ICAR), Greenville, SC 29607, USA}


\maketitle





\begin{abstract}

    Verification and validation (V\&V) of autonomous vehicles (AVs) is critical to ensure operational safety, reliability, and regulatory compliance. This typically requires exhaustive testing across a variety of operating environments and driving scenarios including rare, extreme, or hazardous situations that might be difficult or impossible to capture in reality. Additionally, physical V\&V methods such as track-based evaluations or public-road testing are often constrained by time, financial resources, and inherent safety issues, which motivates the need for virtual proving grounds. However, the fidelity and scalability of simulation-based V\&V methods can quickly turn into a bottleneck, given the sheer amount of test cases that need to be executed. In such a milieu, this work proposes a virtual proving ground that flexibly scales digital twin simulations within high-performance computing clusters (HPCCs) and automates the V\&V process. Here, digital twins enable the creation of high-fidelity virtual representations of the AV and its operating environments, allowing extensive scenario-based testing in precisely controlled yet realistic simulations. Meanwhile, cloud-based HPCC infrastructure brings substantial advantages in terms of computational power and scalability, enabling rapid iterations of simulations, processing and storage of massive amounts of data, and deployment of large-scale test campaigns, thereby reducing the time and cost associated with the V\&V process. We demonstrate the efficacy of this approach through a case study that focuses on the variability analysis of a candidate autonomy algorithm to identify potential vulnerabilities in the perception, planning, and control sub-systems of an AV. The modularity, scalability, and interoperability of the proposed framework are demonstrated by deploying a test campaign comprising 256 test cases on two different HPCC architectures with distinct job schedulers to ensure continuous operation in a publicly shared resource setting. The findings highlight the ability of the proposed framework to accelerate and streamline the V\&V process, thereby significantly compressing ($\sim30\times$) the timeline between AV development and deployment.%

\end{abstract}







\section{Introduction}
\label{Section: Introduction}

\begin{figure}[t]
    \centering
    \includegraphics[width=\linewidth]{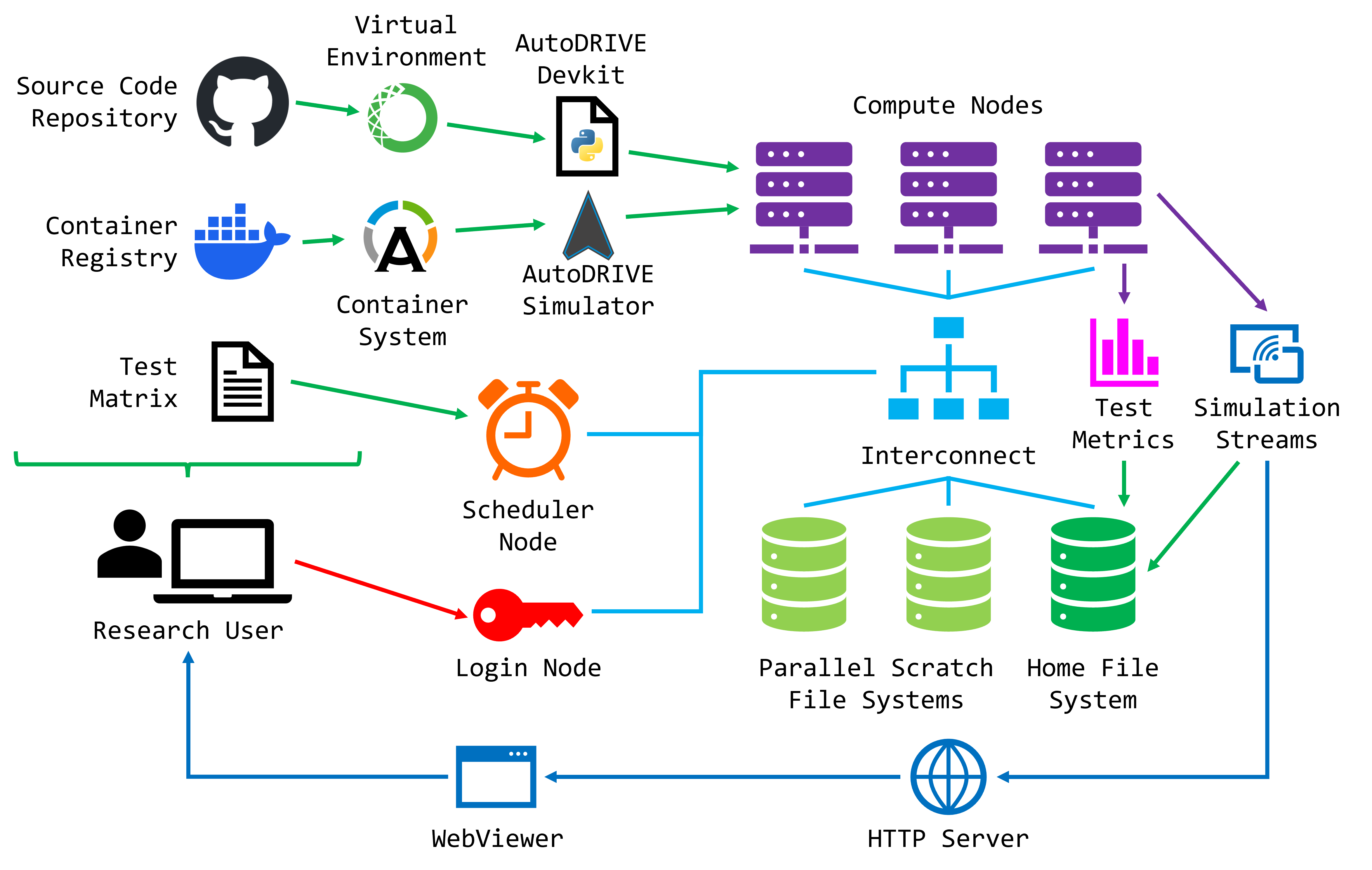}
    \caption{Proposed HPC framework for scalable deployment of digital twin simulations in the cloud.}
    \label{fig1}
\end{figure}

\begin{figure*}[t]
    \centering
    \includegraphics[width=\linewidth]{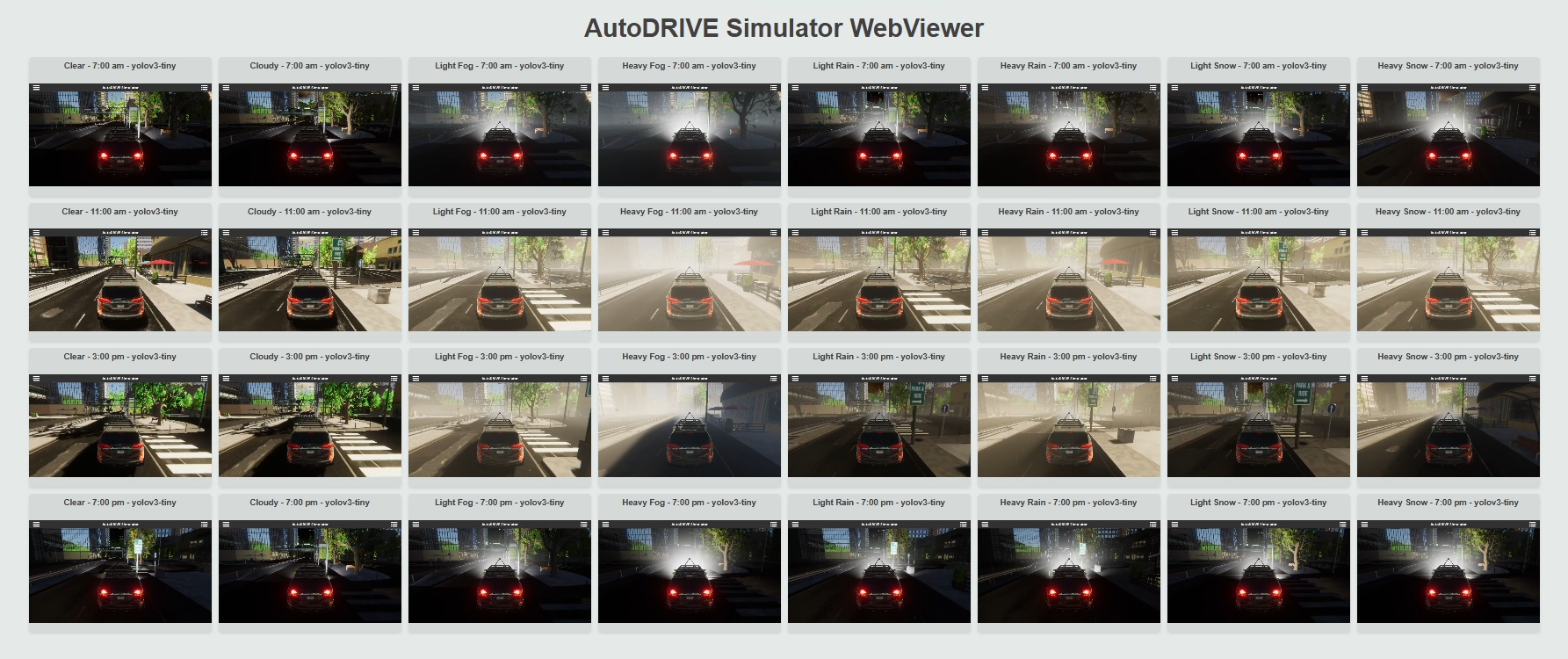}
    \caption{32 parallel simulation instances running within HPCC, visualized on a thin client via a web application.}
    \label{fig2}
\end{figure*}

The rapid advancement of autonomous vehicle (AV) technology presents significant opportunities for improving transportation efficiency and road safety. However, the successful deployment of autonomous systems relies heavily on robust verification and validation (V\&V) processes \cite{10.1145/3542945, 7795548} to ensure that these systems are safe and reliable across a broad spectrum of operating environments and driving scenarios. Automotive testing has traditionally involved track-based (a.k.a. proving grounds) standard benchmark tests as well as lifecycle evaluations based on public road testing in different regions and conditions. While these methods provide valuable insights, they are often constrained by several limitations \cite{75eb35f0-ef72-3076-a9bc-01464345aecc}, including time, cost, and the difficulty of replicating rare, extreme, or hazardous driving scenarios that can pose serious challenges and safety risks \cite{7823109}. This motivates the need for virtual proving grounds \cite{9605690}, which provide the powerful ability to simulate a vast array of driving scenarios in a controlled and risk-free cyber environment. However, simulation fidelity is one of the key concerns in virtual testing. Additionally, given the number of test cases that need to be executed to achieve any statistical significance, the scalability of simulation-based V\&V methods can quickly turn into a bottleneck.

Our work seeks to address these limitations by leveraging recent advancements in digital twin technology, cloud computing, and high-performance computing (HPC). Particularly, this research proposes a modular, scalable, and interoperable framework (refer Fig. \ref{fig1}) to orchestrate digital twin simulations within cloud-based HPC clusters (HPCCs). Here, digital twins provide physically and graphically accurate virtual replicas of both the vehicle and its operating environments, which act as a reliable source of truth for simulation-based V\&V. Coupled with the computational power and scalability of cloud-based HPCC infrastructure, digital twins enable the rapid iteration and large-scale deployment of simulations, significantly reducing the time and cost associated with traditional V\&V methods (refer Fig. \ref{fig2}). This enables continuous, parallelized testing across a wide variety of driving scenes and scenarios, facilitating repeatable and comprehensive performance evaluations. In addition, the proposed framework allows for the immediate adaptation of test conditions, overcoming the bottlenecks inherent in traditional testing setups.

We demonstrate the effectiveness of the proposed framework by presenting a case study that focuses on the variability analysis of a candidate autonomy algorithm. This case study aims to identify potential vulnerabilities in the AV’s perception, planning, and control sub-systems by analyzing several key performance indicators (KPIs). Moreover, we evaluate the performance of the HPCC infrastructure in managing parallel simulation workloads, providing insights into the scalability and efficiency of the framework for large-scale test campaigns. The findings highlight the ability of the proposed framework to accelerate and streamline the V\&V process, thereby significantly compressing the timeline between AV development and deployment.

The key contributions of this work are summarized below:

\begin{itemize}
    \item Developing physically accurate and photorealistic digital twins of a full-scale autonomous vehicle available on campus and its high-fidelity, feature-rich environment, along with a real-time application programming interface (API).
    \item Establishing a modular, scalable, and interoperable HPC framework for continuous, parallelized orchestration of containerized simulation instances in the cluster, while enabling live visualization and interaction from a thin client.
    \item Demonstrating the efficacy of the proposed approach by devising a case study that focuses on the variability analysis of a candidate autonomy algorithm across 256 test cases run on 2 different HPC cluster architectures and job schedulers.
\end{itemize}

The remainder of this paper is structured as follows: Sect.~\ref{Section: Related Work} provides an overview of the relevant literature. Sect.~\ref{Section: Digital Twin Framework} focuses on the modeling and simulation of vehicle and environment digital twins. Sect.~\ref{Section: HPC Framework} explains the configuration and orchestration of parallel simulation workloads within two distinct HPCC architectures. Sect.~\ref{Section: Results and Discussion} presents a systematic variability analysis of the candidate autonomy algorithm along with computational performance assessments of the HPCC. Lastly, Sect.~\ref{Section: Conclusion} concludes the paper and suggests possible avenues for future research.


\section{Related Work}
\label{Section: Related Work}

In this section, we discuss the existing literature relevant to this work by systematically organizing it. We begin by outlining various state-of-the-art simulation frameworks used for simulation-based design (SBD) as well as verification and validation (V\&V) of autonomous vehicles. We then delve into previous research on high-performance computing (HPC), along with the use of simulator containerization and orchestration for cloud deployments. Finally, we highlight some of the prominent research gaps, laying the foundation and motivation for this work.

\subsection{Simulation Frameworks}
\label{Section: Simulation Frameworks}

With the advent of autonomous driving technology, several legacy industrial simulation toolchains have started releasing advanced driver-assistance systems (ADAS) and autonomous driving (AD) features in their updates. Such simulators, including dSPACE ADAS \& AD Portfolio \cite{dSPACE2021}, Hexagon VTD \cite{VTD2025}, Simcenter Prescan \cite{PreScan2025}, Mechanical Simulation CarSim \cite{CarSim2025}, NVIDIA Isaac Sim \cite{IsaacSim2025}, and NVIDIA DRIVE Constellation \cite{DRIVEConstellation2019}, to name a few, offer specialized frameworks to develop and test autonomous driving systems, some of which may be compliant with industrial standards as well. However, the licensing fees and the proprietary nature of such software products can be a limiting factor for the broader community.

The open-source community has also contributed quite a few simulation tools, including Gazebo \cite{Gazebo2004}, CARLA \cite{CARLA2017}, AirSim \cite{AirSim2018}, and LGSVL Simulator \cite{LGSVLSimulator2020}, to name a few. These tools lower the entry barrier by providing free and open-access simulation engines but often struggle with scalability demands. Active maintenance and support also pose a challenge for some open-source projects, especially for smaller developer teams.

\subsection{HPC Frameworks}
\label{Section: HPC Frameworks}

The growing popularity of the software as a service (SaaS) model \cite{tsai2014software} and the widespread use of enterprise-grade platforms like Microsoft Azure, Google Cloud, and Amazon Web Services (AWS) \cite{gupta2021review} allow existing workflows to access substantial computational power. Previous studies have explored the application of HPC resources for modeling and simulation as a service (MSaaS) in different contexts \cite{6721436, 7988851}. These approaches often present significant potential for reducing both time and cost as compared to traditional workflows performing similar tasks \cite{science2019modelling, lyu2020evaluation, franchi2022webots}.

Most of the existing approaches leverage containerization solutions such as Docker \cite{merkel2014docker} and Apptainer (formerly known as Singularity) \cite{singularity2017} to package applications and their dependencies into self-contained units, enabling consistent and portable deployment across different environments. Some approaches \cite{fogli2023chaos, rehman2019cloud, RZR-DT-HPC-2024, 8990167} utilize container orchestration frameworks, such as Kubernetes \cite{Kubernetes2014}, for automating, scaling, and managing the cloud deployments. Additionally, realizing the benefits of HPC infrastructure, several industrial toolchains \cite{VTDx, SIMPHERA, ForetifyV-Suites} have started offering cloud-native services for simulating autonomous driving systems.

\subsection{Research Gaps}
\label{Section: Research Gaps}

Following are a few shortcomings of the existing simulation frameworks, which are addressed by this work.

\begin{itemize}
    \item Some of the existing simulation tools prioritize graphical fidelity at the expense of dynamical accuracy, while others prioritize physical precision over photorealism. Some tools also lose real-time performance due to focusing too intensely on either end of the spectrum.
    \item Most open-source simulators model generic vehicles and environments. They lack precise/calibrated representations of real-world assets, which renders them unsuitable for ``digital twinning'' applications.
\end{itemize}

This work leverages AutoDRIVE Ecosystem\footnote{AutoDRIVE: \url{https://autodrive-ecosystem.github.io}} \cite{AutoDRIVEEcosystem2023, AutoDRIVESimulator2021}, an autonomy-oriented digital twinning ecosystem centered around seamless reality-to-simulation (real2sim) and simulation-to-reality (sim2real) workflows. AutoDRIVE provides physically and graphically accurate digital twins of autonomous vehicles and environments spanning across different scales, configurations, and operational design domains. Additionally, being an open-source ecosystem, it allows users to customize the simulation engine to better fit their use cases.

Following are a few shortcomings of the existing HPC frameworks, which are addressed by this work.

\begin{itemize}
    \item Prior research has typically been highly specialized concerning the choice of individual tools as well as the overall HPC framework. Such systems may not be expandable or interoperable. This can also lead to other issues such as dependency deprecation or vendor lock-in.
    \item There exists a gap in terms of seamlessly deploying existing applications, which were not designed for cloud-based environments, into HPC infrastructure. This complicates the integration of legacy applications into cloud-based HPCCs, necessitating more adaptable and open solutions.
\end{itemize}

This work leverages Palmetto Cluster\footnote{Palmetto: \url{https://docs.rcd.clemson.edu/palmetto}} \cite{Palmetto2024} to propose an open\footnote{GitHub: \url{https://github.com/AutoDRIVE-Ecosystem/AutoDRIVE-HPC}}, modular, and scalable HPC framework for autonomous driving simulations, designed around commonly used platforms, and suitable for a broad range of applications. The proposed framework is agnostic to any specific HPCC architecture and supports containerization, virtual environments, and other workflows to deploy cloud-native as well as legacy applications in cloud-based HPC infrastructure. The proposed framework flexibly allows manual, semi-automated, and fully automated deployment of workloads in the cloud.


\section{Digital Twin Framework}
\label{Section: Digital Twin Framework}

\begin{figure*}[t]
     \centering
     \begin{subfigure}[b]{0.31\linewidth}
         \centering
         \includegraphics[height=0.125\textheight, keepaspectratio]{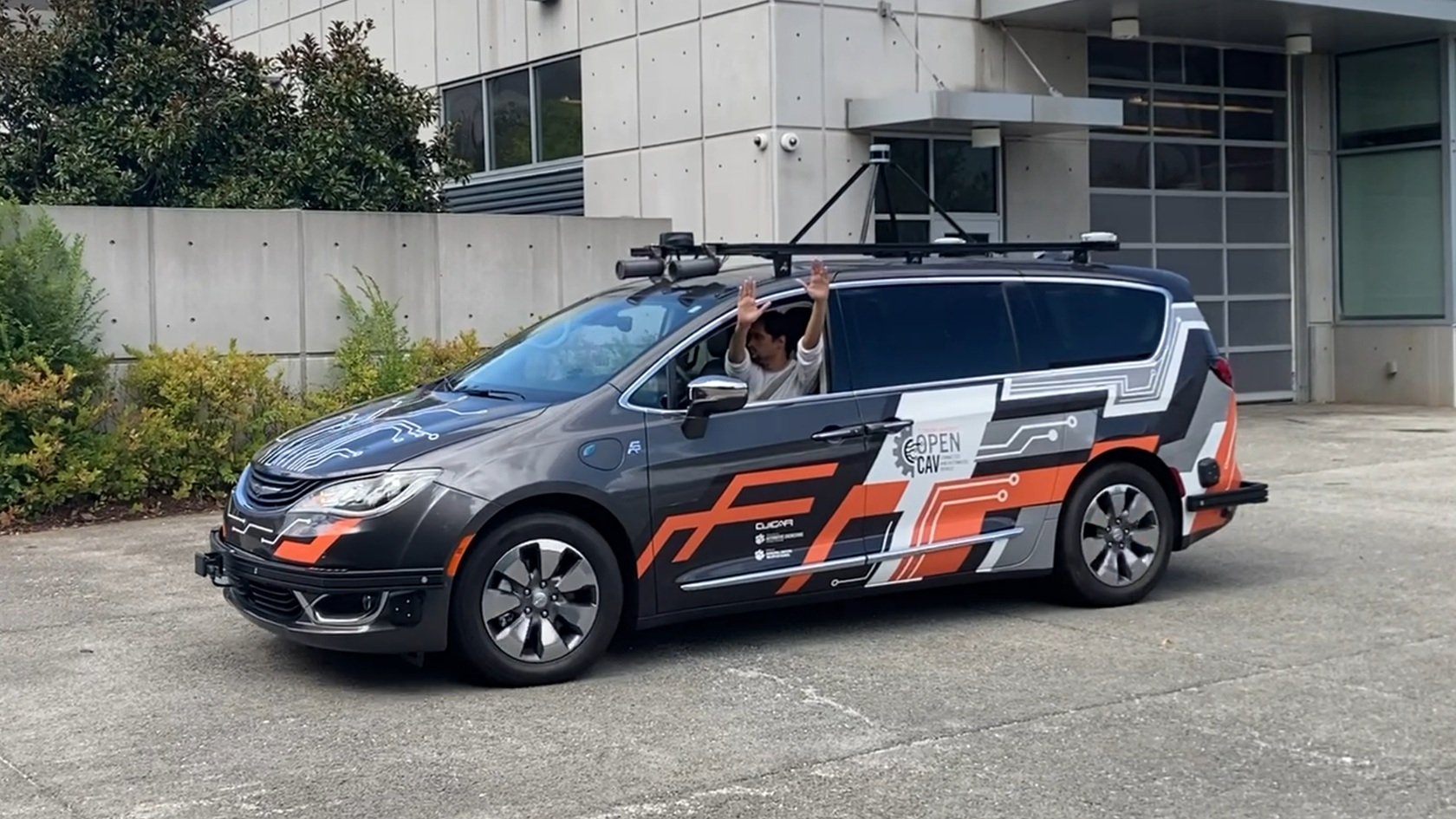}
         \caption{}
         \label{fig3a}
     \end{subfigure}
     \begin{subfigure}[b]{0.31\linewidth}
         \centering
         \includegraphics[height=0.125\textheight, keepaspectratio]{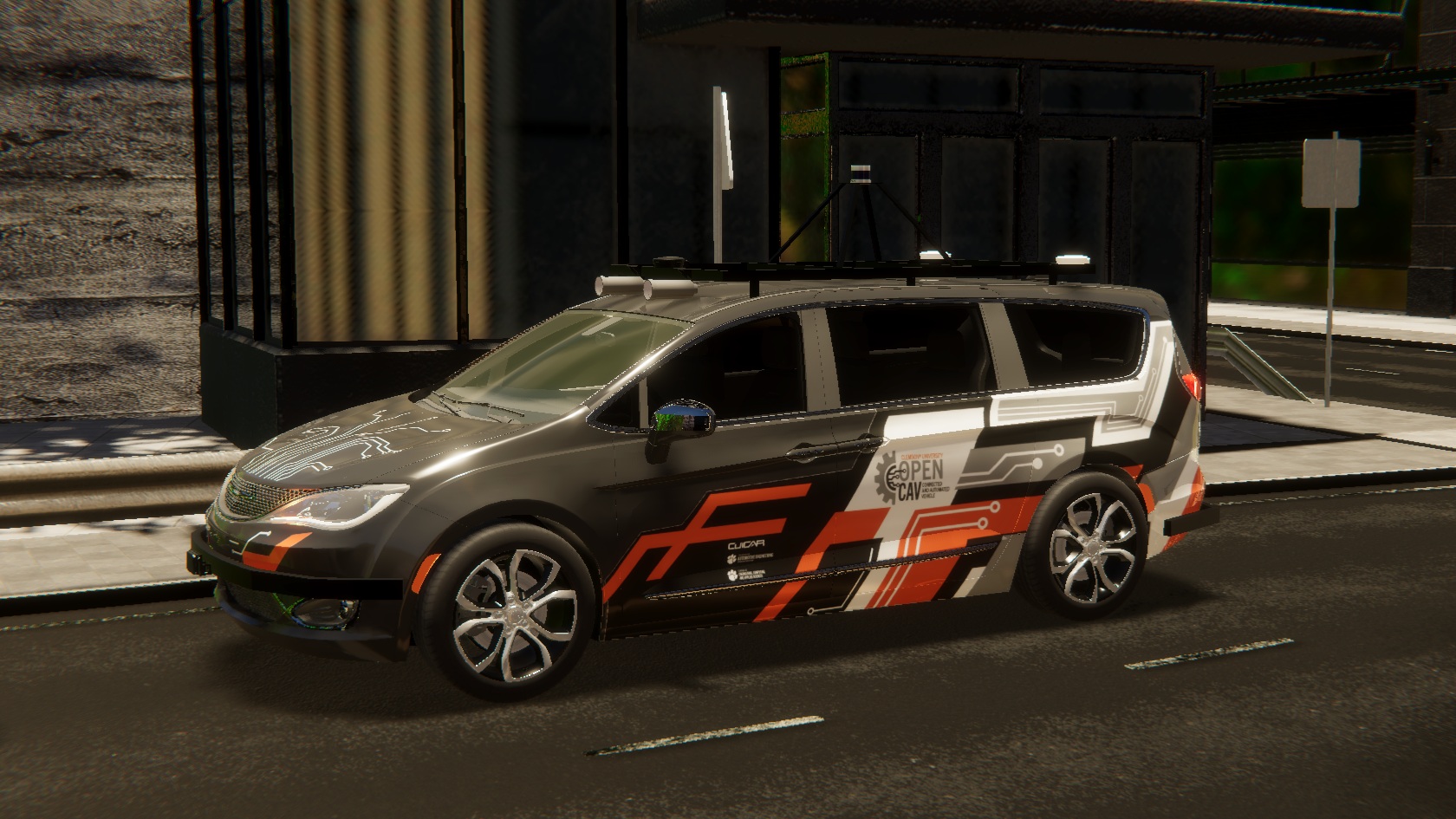}
         \caption{}
         \label{fig3b}
     \end{subfigure}
     \begin{subfigure}[b]{0.37\linewidth}
         \centering
         \includegraphics[height=0.125\textheight, keepaspectratio]{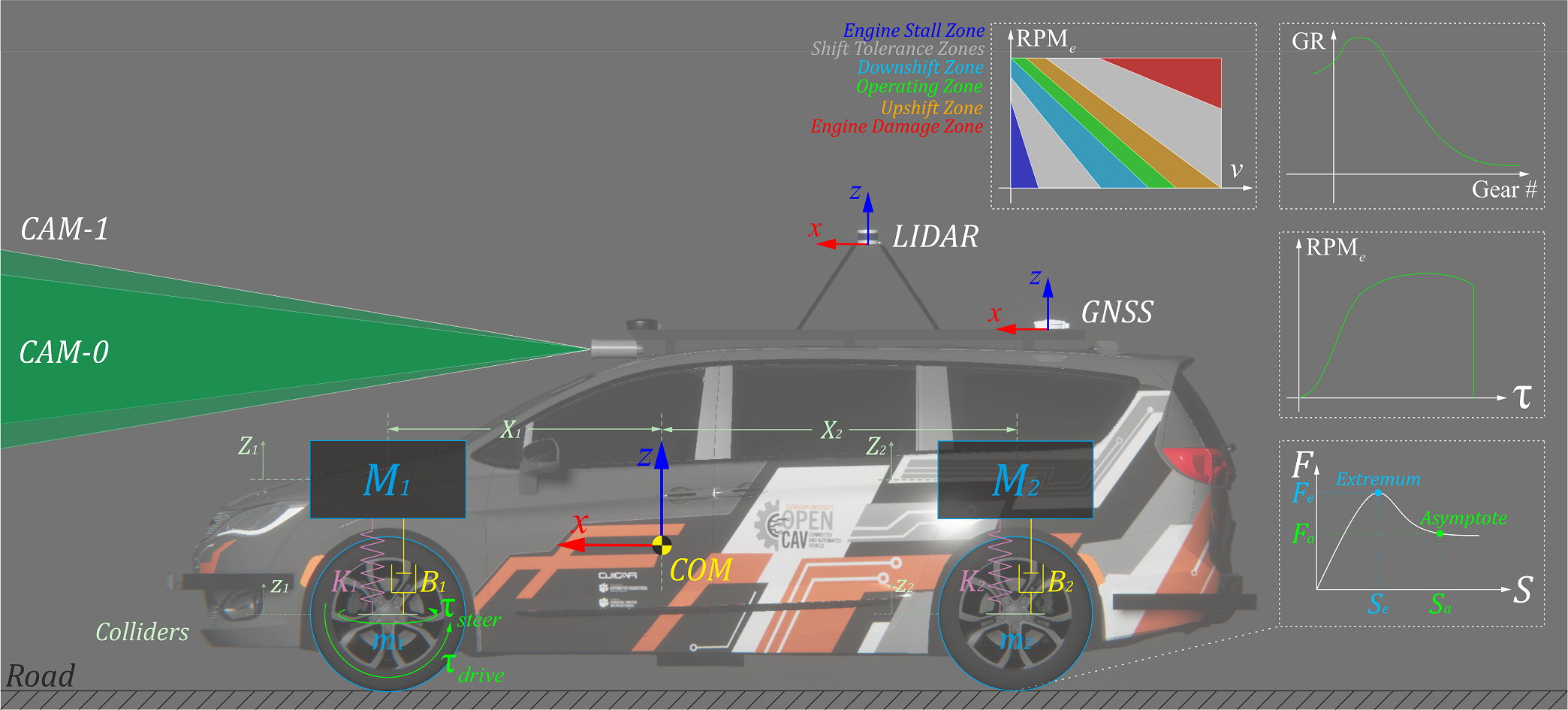}
         \caption{}
         \label{fig3c}
     \end{subfigure}
     \caption{Autonomy-oriented digital twin of OpenCAV: (a) physical twin of OpenCAV at CU-ICAR, (b) digital twin of OpenCAV in AutoDRIVE Simulator, and (c) simplified representation of the vehicle dynamics and sensor simulation.}
    \label{fig3}
\end{figure*}

This section elucidates the creation of high-fidelity digital twins of the OpenCAV\footnote{OpenCAV: \url{https://sites.google.com/view/opencav}} and one of its operating environments within the AutoDRIVE Simulator. Fig. \ref{fig3} depicts the physical (Fig. \ref{fig3a}) and digital (Fig. \ref{fig3b}) twins of the vehicle. The following sections describe vehicle dynamics and powertrain modeling, sensor physics for proprioceptive and exteroceptive perception, as well as static and dynamic environment simulation.

\subsection{Vehicle Simulation}
\label{Section: Vehicle Simulation}

AutoDRIVE Simulator models virtual vehicles (refer Fig. \ref{fig3c}) using a sprung-mass representation ${^iM}$ with rigid-body dynamics. The total mass $M=\sum{^iM}$, center of mass $X_{\text{COM}} = \frac{\sum{{^iM}*{^iX}}}{\sum{^iM}}$ and moment of inertia $I_{\text{COM}} = \sum{{^iM}*{^iX^2}}$ parameters link the two representations cohesively. All the wheels of the vehicle are modeled as independent rigid bodies with mass $m$, and are acted upon by gravitational and suspension forces: ${^im} * {^i{\ddot{z}}} + {^iB} * ({^i{\dot{z}}}-{^i{\dot{Z}}}) + {^iK} * ({^i{z}}-{^i{Z}})$. Here, the the sprung mass ${^iM}$, natural frequency ${^i\omega_n}$, and damping ratio ${^i\zeta}$ determine the stiffness ${^iK} = {^iM} * {^i\omega_n}^2$ and damping $^iB = 2 * ^i\zeta * \sqrt{{^iK} * {^iM}}$ of the suspension system.

Powertrain of the vehicle is modeled to produce a torque $\tau_{\text{total}} = \left[\tau_e\right]_{RPM_e} * \left[GR\right]_{G_\#} * FDR * \tau * \mathscr{A}$, were, $\tau$ is the throttle input, $\tau_e$ is the engine torque, and $\mathscr{A}$ is a non-linear smoothing operator. Refer to the top-right and mid-right insets in Fig. \ref{fig3c} for the $\left[\tau_e\right]_{RPM_e}$ and $\left[GR\right]_{G_\#}$ mappings. The engine speed is updated as $RPM_e := \left[RPM_i + \left(|RPM_w| * FDR * GR\right)\right]_{(RPM_e,v)}$ where, $RPM_i$ is the engine idle speed, $RPM_w$ is the average wheel speed, $FDR$ is the final drive ratio, $GR$ is the gear ratio, and $v$ is the body velocity. The automatic transmission keeps the engine speed in check by maintaining a good operating range (refer to the top-left inset in Fig. \ref{fig3c}): $RPM_e = \frac{{v_{\text{MPH}} * 5280 * 12}}{{60 * 2 * \pi * R_{\text{tire}}}} * FDR * GR$. The torque at the differential depends on the drivetrain configuration:
$
\tau_{\text{diff}} = \begin{cases}
\frac{\tau_{\text{total}}}{2} & \text{if FWD/RWD} \\
\frac{\tau_{\text{total}}}{4} & \text{if AWD}
\end{cases}
$.
The differential is modeled to divide this torque to the left $^{L}\tau_{w} = \tau_{\text{diff}} * (1 - \tau_{\text{drop}} * |\delta^{-}|)$ and right $^{R}\tau_{w} = \tau_{\text{diff}} * (1 - \tau_{\text{drop}} * |\delta^{+}|)$ wheels based on the steering input $\delta$. Here, $\tau_w$ is the torque at wheels and $\tau_{\text{drop}}$ is the torque drop at the differential. The value of $(\tau_{\text{drop}} * |\delta^{\pm}|)$ is clamped between $[0,0.9]$.

Braking sub-system of the vehicle is modeled to produce a torque ${^i\tau_{\text{brake}}} = \frac{{^iM}*v^2}{2*D_{\text{brake}}}*R_b$, where $D_{\text{brake}}$ is the braking distance (calibrated at 60 MPH) and $R_b$ is the brake disk radius.

The steering sub-system of the vehicle is modeled based on the Ackermann steering geometry. The kinematics of this mechanism are governed by the wheelbase $l$ and the track width $w$, expressed as
$ \delta_{l/r} = \textup{tan}^{-1}\left(\frac{2*l*\textup{tan}(\delta)}{2*l\pm w*\textup{tan}(\delta)}\right)$.
Dynamically, the steering rate $\dot{\delta} = \kappa_\delta + \kappa_v * \frac{v}{v_{\text{max}}}$ is governed by the steering sensitivity $\kappa_\delta$ and the speed-dependency factor $\kappa_v$ of the steering mechanism.

The tires of the vehicle are modeled using non-linear friction curves represented as $\left\{\begin{matrix} {^iF_{t_x}} = F(^iS_x) \\{^iF_{t_y}} = F(^iS_y) \\ \end{matrix}\right.$, where $^iS_x$ and $^iS_y$ denote the longitudinal and lateral slip values of the $i$-th tire. Here, each friction curve is modeled using a two-piece cubic spline $F(S) = \left\{\begin{matrix} f_0(S); \;\; S_0 \leq S < S_e \\ f_1(S); \;\; S_e \leq S < S_a \\ \end{matrix}\right.$, where $f_k(S) = a_k*S^3+b_k*S^2+c_k*S+d_k$. The first segment of the said spline spans from the origin $(S_0,F_0)$ to the extremum $(S_e,F_e)$ point, and the second segment spans from the extremum $(S_e, F_e)$ point to the asymptote $(S_a, F_a)$ point (refer to the bottom-right inset in Fig. \ref{fig3c}).

The vehicle aerodynamics accounts for variable air drag, based on the operating condition:

$ F_{\text{aero}} = 
\begin{cases}
F_{d_{\text{max}}} & \text{if } v \geq v_{\text{max}} \\
F_{d_{\text{idle}}} & \text{if } \tau_{\text{out}} = 0 \\
F_{d_{\text{rev}}} & \text{if } (v \geq v_{\text{rev}}) \land (G_\# = -1) \land (RPM_{w} < 0) \\
F_{d_{\text{run}}} & \text{otherwise}
\end{cases}
$
Here, $v$ is the vehicle velocity, $v_{\text{max}}$ is the maximum practical forward velocity, $v_{\text{rev}}$ is the maximum practical reverse velocity, $G_\#$ is the operating gear, and $RPM_w$ is the average wheel speed.

\subsection{Sensor Simulation}
\label{Section: Sensor Simulation}

\begin{figure*}[t]
     \centering
     \begin{subfigure}[b]{0.49\linewidth}
         \centering
         \includegraphics[width=\linewidth]{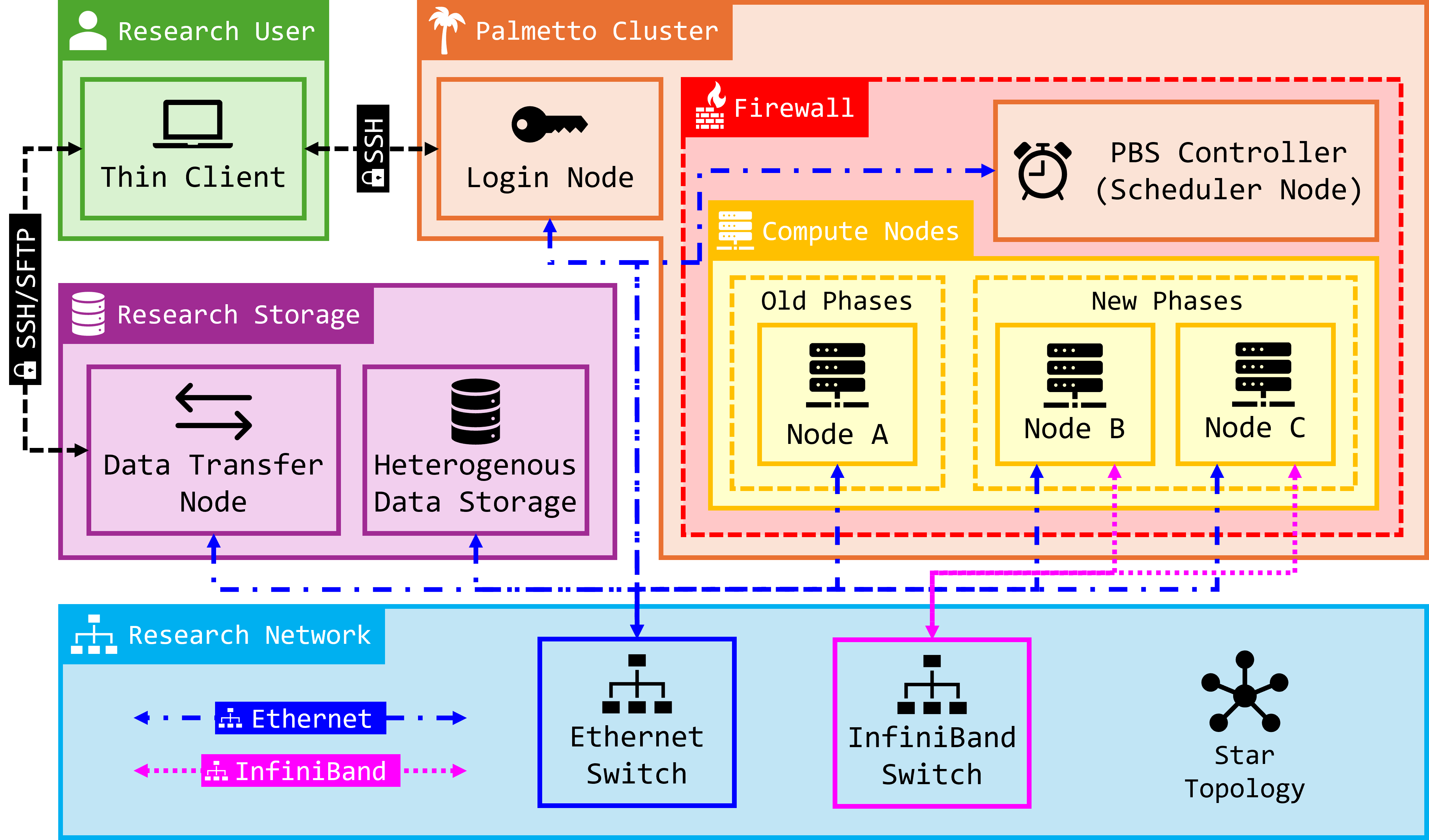}
         \caption{}
         \label{fig4a}
     \end{subfigure}
     \hfill
     \begin{subfigure}[b]{0.49\linewidth}
         \centering
         \includegraphics[width=\linewidth]{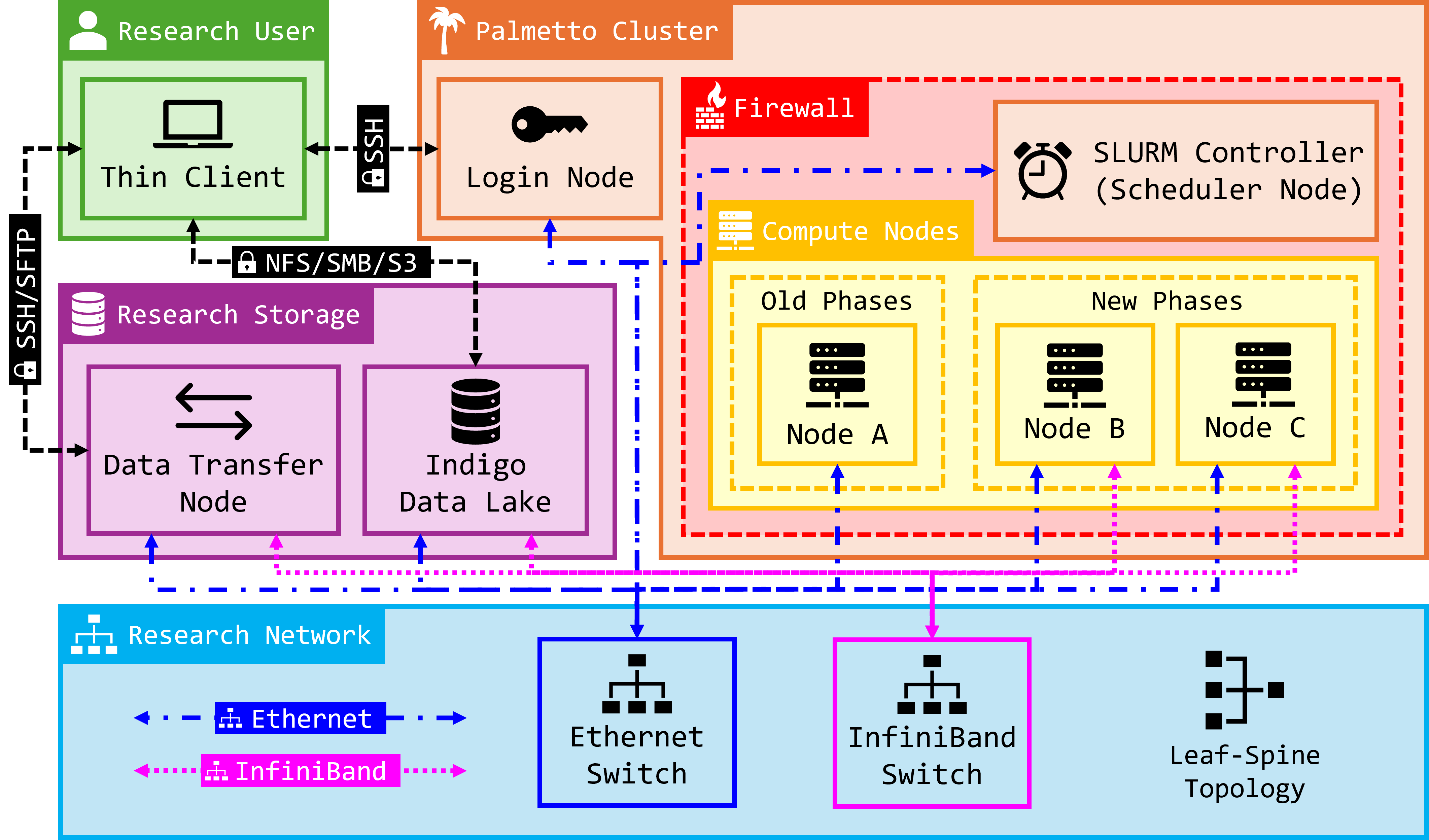}
         \caption{}
         \label{fig4b}
     \end{subfigure}
     \caption{High-level architecture of (a) Palmetto 1 Cluster with PBS and (b) Palmetto 2 Cluster with SLURM.}
    \label{fig4}
\end{figure*}

The autonomy-oriented digital twin of OpenCAV is equipped with a range of virtual sensors, with modulatable noise injection and compensation via APIs. These sensors are modeled and calibrated using physics-based computational techniques to faithfully replicate the characteristics of their real-world counterparts onboard the physical OpenCAV.

The simulated proprioceptive sensors include by-wire actuator feedbacks, incremental encoders, a global navigation satellite system (GNSS), and an inertial measurement unit (IMU). The simulator provides real-time feedback from the throttle ($\tau$), steering ($\delta$), brake ($\chi$), and parking brake ($\xi$) of the vehicle. The virtual encoders measure incremental displacement of the wheels, $^iN_{\text{ticks}} = {^iPPR} * {^iCGR} * {^iN_{\text{rev}}}$, where $^iN_{\text{rev}}$ denotes the wheel revolutions, $^iPPR$ denotes the encoder resolution, and $^iCGR$ is the cumulative gear ratio. The virtual inertial navigation system (INS) simulates GNSS and IMU based on temporally coherent rigid-body transform updates of the vehicle $\{v\}$ w.r.t. the world $\{w\}$: ${^w\mathbf{T}_v} = \left[\begin{array}{c | c} \mathbf{R}_{3 \times 3} & \mathbf{t}_{3 \times 1} \\ \hline \mathbf{0}_{1 \times 3} & 1 \end{array}\right] \in SE(3)$.

The simulated exteroceptive sensors include 2 RGB cameras and a 3D LIDAR. The virtual cameras are modeled using a standard viewport rendering pipeline, which starts by computing the camera view matrix, denoted as $\mathbf{V} \in SE(3)$, based on the relative homogeneous transformation of the camera $\{c\}$ w.r.t. the world $\{w\}$. This is followed by computing the camera projection matrix $\mathbf{P} \in \mathbb{R}^{4 \times 4}$ to map world coordinates onto the image space. The projection matrix is given by:
$\mathbf{P} = \begin{bmatrix} \frac{2*N}{R-L} & 0 & \frac{R+L}{R-L} & 0 \\ 0 & \frac{2*N}{T-B} & \frac{T+B}{T-B} & 0 \\ 0 & 0 & -\frac{F+N}{F-N} & -\frac{2*F*N}{F-N} \\ 0 & 0 & -1 & 0 \\ \end{bmatrix}$
where $N$ and $F$ represent the near and far clipping planes of the camera, respectively, while $L$, $R$, $T$, and $B$ correspond to the left, right, top, and bottom sensor offsets. Finally, a post-processing pipeline is adopted to simulate various physical effects such as lens distortion, depth of field, exposure, ambient occlusion, contact shadows, bloom, motion blur, film grain, chromatic aberration, and other similar lens and film-related effects.

The virtual 3D LIDAR is simulated through parallelized multi-channel ray-casting with GPU acceleration (if available): \texttt{raycast}\{$^w\mathbf{T}_l$, $\vec{\mathbf{R}}$, $r_{\text{max}}$\} for each angle $\theta \in \left [ \theta_{\text{min}}:\theta_{\text{res}}:\theta_{\text{max}} \right ]$ and each channel $\phi \in \left [ \phi_{\text{min}}:\phi_{\text{res}}:\phi_{\text{max}} \right ]$, at a specific update rate. Here, $r_{\text{min}}$ and $r_{\text{max}}$, $\theta_{\text{min}}$ and $\theta_{\text{max}}$, and $\phi_{\text{min}}$ and $\phi_{\text{max}}$ respectively denote the minimum and maximum linear ranges, horizontal angular ranges, and vertical angular ranges, while $\theta_{\text{res}}$ and $\phi_{\text{res}}$ represent the horizontal and vertical angular resolutions of the LIDAR, respectively. In this model, ${^w\mathbf{T}_l} = {^w\mathbf{T}_v} * {^v\mathbf{T}_l} \in SE(3)$ denotes the relative transformation of the LIDAR \{$l$\} w.r.t. the vehicle \{$v$\} and the world \{$w$\}. The direction vector for each ray-cast $R$ is given by $\vec{\mathbf{R}} = \left [\cos(\theta)*\cos(\phi) \;\; \sin(\theta)*\cos(\phi) \;\; -\sin(\phi) \right ]^T$. The resulting ray-cast hits $\{\mathbf{x_{\text{hit}}, y_{\text{hit}}, z_{\text{hit}}}\}$ are thresholded, encoded, and returned as 3D point cloud data (PCD).

\subsection{Environment Simulation}
\label{Section: Environment Simulation}

AutoDRIVE Simulator offers a range of environment creation options including terrain generation, mesh modeling, and 3D scene reconstruction. The static scenes are simulated through mesh-mesh interference detection, which calculates contact forces, friction, momentum transfer, and air drag on all rigid bodies. Dynamic variability can be introduced by changing the time of day and weather conditions in real time. The simulator models physically-based sky and celestial bodies, allowing precise ray tracing to create real-time or pre-baked light maps. This enables the simulation of horizon gradients, as well as the reflection, refraction, diffusion, scattering, dispersion, and attenuation of light across different materials. The weather phenomena are procedurally generated using physically-based volumetric and particle effects. These include static and dynamic clouds, volumetric fog and mist, precipitation particles (rain and snow), and stochastic wind gusts. The independent weather elements can be altered in real-time or preset conditions can be simulated on demand (e.g., sunny, cloudy, foggy, rainy, snowy, etc.).


\section{HPC Framework}
\label{Section: HPC Framework}

This section presents a high-level architectural overview of the proposed HPC framework along with the hardware-software specifications of the HPCC infrastructure. It further delves into the job array scheduling used for parallelizing workloads, and discusses a flexible deployment approach offering headless (no graphics), record-and-replay (record streams to file), and live-streaming (real-time interactive visualization) variants.

\subsection{Architecture Overview}
\label{Section: Architecture Overview}

The proposed HPC framework (refer Fig. \ref{fig1}) is formulated considering best practices and user convenience. Furthermore, we adopt a modular open system architecture, making it agnostic to any specific component(s) or service(s). Particularly, the research user only needs to define 3 things viz. upstream repository (or container registry) which holds the codebase and dependency list for the autonomy algorithm, container registry (or upstream repository) of the digital twin simulator, and a test matrix configuration file which defines the different test parameters. Once the user logs in (via the login node) and submits a job array with this information, the framework parses the test matrix and invokes the scheduler node to pull the appropriate repositories and container images to the file system, spin up multiple simulator and devkit instances based on available compute nodes, and begin variability testing across the different test parameters. All the generated data and test metrics are logged automatically to the file system, with an option to live-stream all the simulation visualizations to a thin client via a web app hosted by an HTTP server  (refer Fig. \ref{fig2}).

\subsection{Cluster Specifications}
\label{Section: Cluster Specifications}

Palmetto Cluster is the high-performance computing (HPC) resource of Clemson University, designed and deployed by the Clemson Computing and Information Technology (CCIT) Research Computing and Data (RCD) Infrastructure Group. Palmetto has been benchmarked at 3.01 PFlop/s and consistently ranks among the top 500 most powerful computer systems \cite{TOP500}.

Ever since its inception in 2008, Palmetto has been continuously upgraded to support growing research computing demands. However, 2024 marked a radical shift in the cluster architecture \cite{Palmetto2024}, as it was ported from Palmetto 1, which used the Portable Batch System (PBS) scheduler, to Palmetto 2, which employed the Simple Linux Utility for Resource Management (SLURM) scheduler. Fig. \ref{fig4} depicts the high-level architectures of both versions of the Palmetto cluster. Luckily, this research project was initiated in 2024, and we could leverage both architectures to report our findings. Following is a brief overview of the cluster architectures detailed in Fig. \ref{fig4a} and Fig. \ref{fig4b}:

\begin{itemize}
    \item \textbf{Login Nodes:} The login nodes serve as a gateway between Palmetto and external networks. Since compute nodes are not connected to the internet, users must securely log in to access them.
    \item \textbf{Scheduler Node:} The scheduler node is responsible for job scheduling and resource allocation. The scheduler will add job(s) to a queue and determine which nodes can run the prescribed job(s) and when they will be available.
    \item \textbf{Compute Nodes:} The compute nodes are equipped with high-performance hardware to perform fast calculations on large amounts of data. For example, the general access partition of the cluster cumulatively contains 1,158 nodes; 1,592 GPUs; 47,252 CPU cores; and 309,255 GB of memory. The cluster also hosts additional resources, which have been provisioned/prioritized for specific research projects. It is important to note that, unlike many HPCCs, Palmetto is a heterogeneous cluster with varying hardware configurations across compute nodes. However, Palmetto maintains intra-phase homogeneity, ensuring that all compute nodes within a specific phase share the same hardware configuration.
    \item \textbf{Data Storage:} Palmetto 1 provisioned heterogeneous long-term group storage, three different \texttt{/scratch} spaces (1.8 PB, flexible allocation), and a \texttt{/home} directory (100 GB per user). Palmetto 2 utilizes a homogeneous storage solution named Indigo, powered by VAST Data. The \texttt{/home} and \texttt{/scratch} spaces are now incorporated within Indigo and are respectively limited to 250 GB and 5 TB per user. Apart from these, the compute nodes have a \texttt{/local\_scratch} varying from 99 GB to 2.7 TB per node.
    \item \textbf{Data Transfer Nodes:} The data transfer nodes allow users to move large amounts of data in or out of the cluster. Palmetto 1 primarily supported SFTP, with restrictions on moving data outside of the local Palmetto network. Palmetto 2 rectified this by focusing on data portability -- Indigo is accessible by all entities in the cluster as well as user workstations on the campus network and supports multi-protocol (e.g., NFS, SMB, S3) file sharing.
    \item \textbf{Interconnects:} All the nodes in the cluster are connected via computer network connections, called interconnects. Palmetto supports 2 types of interconnects: Ethernet and InfiniBand. The former is a slower interconnect, but it is available on every node. It supports 1, 10, and 25 Gbps bandwidths. The latter is a specialized, low-latency interconnect offering FDR (56 Gbps), EDR (100 Gbps), and HDR (200 Gbps) bandwidths.
\end{itemize}

\subsection{Parallel Simulation Workloads}
\label{Section: Parallel Simulation Workloads}

Initial simulation workloads were submitted as interactive jobs to allow debugging and troubleshooting while setting up the framework. However, the broader goal of this project was enabling scripted batch jobs to automate the orchestration of parallel simulation instances, thereby harnessing the true power of HPC. This goal was systematically achieved by devising 3 different deployment configurations:

\begin{itemize}
    \item \textbf{Headless Workloads:} This deployment hosts AutoDRIVE Simulator headlessly (rendering disabled) on the Palmetto Cluster. This deployment allows users to ensure that the different software components are configured correctly and communicating with each other. This can be useful for early-stage prototyping, as in our case, or faster execution for applications that do not require graphics rendering.
    \item \textbf{Record-and-Replay Workloads:} This deployment implements an X virtual framebuffer (Xvfb) to perform graphical operations of AutoDRIVE Simulator in virtual memory on the Palmetto Cluster. The virtual renderings are exported as video files, which can be replayed later on demand. This deployment allows users to ensure that the simulator can access the GPU via the prescribed graphics API. This can be useful for late-stage prototyping, as in our case, or unmonitored execution with an option of on-demand replay.
    \item \textbf{Live-Streaming Workloads:} This deployment also leverages Xvfb to perform graphical operations of AutoDRIVE Simulator in virtual memory on the Palmetto Cluster. However, the virtual renderings are exported as HTTP live streams (HLS), which can be visualized live on a thin client connected to the cluster. This is the final deployment configuration, which allows live monitoring of the test cases via an interactive web application\footnote{Video: \url{https://youtu.be/HaOHLP7-9co}} (refer Fig. \ref{fig2}).
\end{itemize}


\section{Results and Discussion}
\label{Section: Results and Discussion}

\begin{figure*}[t]
    \centering
    \includegraphics[width=\linewidth]{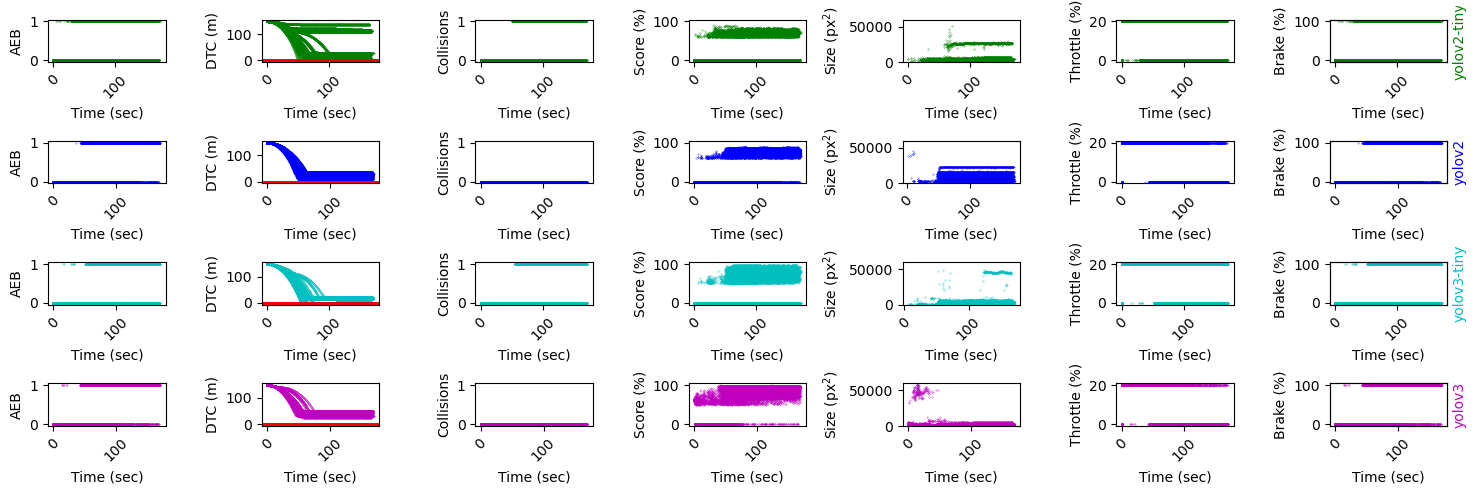}
    \caption{Detailed KPI analysis of the candidate autonomy algorithm swept across a battery of 256 test cases.}
    \label{fig5}
\end{figure*}

This section demonstrates the effectiveness of the proposed framework through a case study centered around the variability analysis of an autonomy algorithm. It also assesses the capability of the HPCC infrastructure in handling parallel simulation workloads, offering insights into the scalability and performance of the framework for large-scale testing.

\subsection{Case Study}
\label{Section: Case Study}

We devised an intuitive autonomous driving case study to demonstrate the efficacy of the proposed framework. Particularly, a jump-scare scenario was laid out to evaluate the reaction of the ego vehicle to a sudden "panic" event. This test scenario dictated that an autonomous vehicle would navigate an urban downtown using visual servoing and keep driving unless doing so became unsafe or unethical (e.g., due to a dead-end, imminent collision, etc.). Further along the road, the ego vehicle would encounter a broken-down vehicle obstructing its lane, prompting the ego to initiate an emergency braking maneuver.

We implemented a 3-stage reactive algorithm to fit this use case. First, the perception sub-system would acquire an RGB frame from the front-right camera of the ego vehicle, resize it, and pass it through an object detection model. The detections would then be fed to a finite-state machine for planning the reactive behavior. The planner accounted for the class, confidence, and size of the detections to generate an autonomous emergency braking (AEB) trigger. The controller then regulates the drive-by-wire system of the ego vehicle to realize the emergency braking maneuver when required. Furthermore, since the primary autonomy algorithm utilized visual perception, a secondary algorithm was implemented to adjust the vehicle lights (off, low-beam, high-beam, fog-lights) according to the ambient light and the presence of fog or mist. These values were determined by considering the time of day and current weather conditions.

We analyzed the performance of 4 different state-of-the-art deep-learning models for object detection. These included \texttt{yolov2}, \texttt{yolov2-tiny}, \texttt{yolov3}, and \texttt{yolov3-tiny}. It is to be noted that these models were deployed ``as is'' without any re-training, and serve as mere examples of algorithmic variants. Similarly, the end-to-end AEB algorithm described in this work is also intended to serve as an example. The autonomy algorithm itself is beyond the scope of this study. This work primarily focuses on the novel HPC framework, which is completely modular and open-source, thereby readily accommodating any autonomy algorithm.

\subsection{Autonomy V\&V KPI Analysis}
\label{Section: Autonomy V&V  KPI Analysis}

The overall AEB scenario was divided into several test cases to evaluate the performance of the system under test (SUT) with 4 candidate perception models\{\texttt{yolov2}, \texttt{yolov2-tiny}, \texttt{yolov3}, \texttt{yolov3-tiny}\} across 8 different times of day \{5 am, 7 am, 9 am, 11 am, 1 pm, 3 pm, 5 pm, 7 pm\}, and 8 different weather conditions \{clear, cloudy, light fog, heavy fog, light rain, heavy rain, light snow, heavy snow\}. These rather sparse parameter sets quickly resulted in a significantly large test matrix, which amounted to a total of 256 test cases. These cases were grouped into 8 batches, with each batch running 32 test cases, covering all the combinations. It is important to note that, depending on the specific SUT and verification requirements, the total number of test cases can be derived from the test matrix, and simulation instances can be executed accordingly to conduct a batched variability analysis of the algorithm.

We identify a test case instance as ``successful'' if there haven’t been any collisions. With this simple binary metric, we obtain the following overall results. The high-level ``pass/fail'' outcomes from the variability analysis (refer Table \ref{tab1}) reveal that \texttt{yolov2} and \texttt{yolov3} are the most effective perception models with a success rate of 100\%. However, looking at the compute resource utilization (refer Table \ref{tab2}), we can infer that in this case, \texttt{yolov2} could be the better choice since it offers similar performance at a lower computational expense. \texttt{yolov2-tiny} with a success rate of only 42.19\%, performs the worst, even below \texttt{yolov3-tiny} reporting 85.94\% success. A deeper analysis of the KPIs offers valuable insights into potential weaknesses in the perception, planning, and control modules of the autonomy stack, which can aid in identifying and addressing issues through algorithm improvements (refer to Fig. \ref{fig5}).

Particularly interesting KPIs are the first column, which plots the AEB trigger binary value, and the second column, which shows the distance to collision (DTC). In most simulation instances, we observe that DTC decreases down to a small non-zero value, indicating that the AEB trigger was successfully timed. Here, it is also worth mentioning that \texttt{yolov3} maintained the lowest proximity to the factor of safety (FOS) mark (indicated by the red line in Fig. \ref{fig5}), which was set to be 1 m, indicating a highly reliable performance. The third column plots the collision count over time. When the test is unsuccessful (i.e., the ego vehicle is unable to stop in time), the collision count goes to 1. It was interesting to see that the choice of vehicle lighting (headlights vs. fog lights) has a significant impact on the algorithmic performance, especially in rainy and foggy conditions. Finally, the last two columns plot the throttle and braking commands over time. Again, an interesting finding was that the effects of weather on vehicle controls cannot be ignored (e.g. varying traction conditions during rain/snow, making driving and braking challenging).

\begin{table}[t]
\centering
\caption{High-level outcomes of autonomy V\&V}
\resizebox{\columnwidth}{!}{%
\begin{tabular}{llll}
\toprule
\textbf{Batch ID} & \textbf{System Under Test} & \textbf{Test Cases Passed} & \textbf{Total Test Cases} \\
\midrule
\{1, 2\} & \texttt{yolov2}      & 64 & 64 \\
\{3, 4\} & \texttt{yolov2-tiny} & 27 & 64 \\
\{5, 6\} & \texttt{yolov3}      & 64 & 64 \\
\{7, 8\} & \texttt{yolov3-tiny} & 55 & 64 \\
\bottomrule
\textbf{Cumulative} &  & 210 & 256 \\
\bottomrule
\end{tabular}%
}
\label{tab1}
\end{table}

\begin{table}[t]
\centering
\caption{Net compute resource utilization analysis}
\resizebox{\columnwidth}{!}{%
\begin{tabular}{lllll}
\toprule
\textbf{Batch ID} & \textbf{CPU Cores} & \textbf{RAM (GB)} & \textbf{GPU (\%)} & \textbf{GPU Memory (GB)} \\
\midrule
1 & 145 & 126.74 & 32 & 86.4 \\
2 & 152 & 139.21 & 36 & 82.8 \\
3 & 128 & 119.36 & 23 & 74.6 \\
4 & 130 & 117.56 & 26 & 69.9 \\
5 & 180 & 147.91 & 47 & 89.7 \\
6 & 172 & 151.43 & 41 & 91.4 \\
7 & 117 & 123.78 & 31 & 85.2 \\
8 & 122 & 122.69 & 28 & 85.9 \\
\bottomrule
\textbf{Mean} & 143 & 131.09 & 33 & 83.2 \\
\bottomrule
\end{tabular}%
}
\label{tab2}
\end{table}

\subsection{Computational Performance}
\label{Section: Computational Performance}

The computational performance metrics were logged at 0.2 Hz, for each of the batch jobs comprising 32 parallel simulation instances. Table \ref{tab2} hosts the peak resource utilization for each of the 8 batches described earlier. It can be observed that while all the batches are utilizing roughly the same amount of resources, batches 3, 4, 7, and 8 are slightly more performant than others. These are the batches running \texttt{tiny} model variants, which contain a lower number of parameters, and therefore show reduced resource utilization, in exchange for lower detection performance (as evident from the failure cases). This is because, while the simulators are the primary workload in this case, the choice of perception model also has an impact on the resources. This small effect, when aggregated across a batch of 32 instances, can be significant.

From the perspective of the digital twin simulations, we noted frame rates exceeding 30 Hz when the simulator was running at maximum fidelity and full HD (i.e. 1920$\times$1080p), with rates rising above 60 Hz as the simulation fidelity and target resolution were lowered. While the simulation timestep remained independent of the frame rate to maintain physical accuracy, the increase in frame rate contributed to a higher real-time factor for the simulations, improving the overall speed of test-case execution.

From the HPC deployment perspective, the utility of the proposed framework is apparent from the fact that an entire batch of 32 tests was executed within 3 minutes of wall time. Considering a perfectly efficient serial execution of the same tests, it would take over 1.6 hours. This marks a theoretical speed-up of 32$\times$. Considering the practical challenges in job submission and resource allocation, we can still expect an efficiency of over 25$\times$.


\section{Conclusion}
\label{Section: Conclusion}

In conclusion, the proposed virtual proving ground framework, leveraging digital twin simulations and high-performance computing clusters, offers a transformative approach to the verification and validation of autonomous vehicles. By addressing the challenges of scalability, fidelity, and resource constraints, this framework enables extensive scenario-based testing in a virtual environment, thus reducing the need for time-consuming, costly, and risky physical testing. The demonstrated case study highlights the potential of this approach to identify the vulnerabilities in autonomous driving solutions, accelerate the V\&V process, and significantly reduce development timelines. The modular, scalable, and interoperable nature of the framework makes it a promising solution for large-scale, efficient V\&V campaigns, ultimately enhancing the safety, reliability, and regulatory compliance of autonomous vehicles while ensuring faster deployment.

Future research could focus on integrating artificial intelligence (AI) based methods to automate test case generation and anomaly detection. Additionally, improving adaptive cloud infrastructure for resource efficiency, incorporating real-time data from live deployments, and addressing ethical and safety considerations in edge cases could further advance the effectiveness and scope of such virtual proving grounds for autonomous vehicle testing.






\bibliographystyle{asmeconf}  
\bibliography{references}


\end{document}